\documentclass{article}

\usepackage{amsmath}
\usepackage[T1]{fontenc} 
\usepackage{color}
\usepackage{apacite}

\begin{document}

\title{The origins of Zipf's meaning-frequency law}

\author{Ramon Ferrer-i-Cancho$^{1,*}$ and Michael S. Vitevitch$^2$\\
$^1$ {\small Complexity and Quantitative Linguistics Lab. LARCA Research Group. } \\ 
{\small Departament de Ci\`encies de la Computaci\'o, Universitat Polit\`ecnica de Catalunya. } \\
{\small Campus Nord, Edifici Omega, Jordi Girona Salgado 1-3. 08034 Barcelona, Catalonia (Spain). } \\ 
{\small Phone: +34 934134028. E-mail: rferrericancho@cs.upc.edu. } \\
$^2$ {\small Spoken Language Laboratory. Department of Psychology. University of Kansas. } \\ 
{\small 1415 Jayhawk Blvd. Lawrence, KS 66045, USA.} \\ 
{\small Phone: +1 (785) 864-9312. E-mail: mvitevit@ku.edu.} \\
$^*$ {\small Author for correspondence.}
}

\maketitle 

\begin{abstract}
In his pioneering research, G. K. Zipf observed that more frequent words tend to have more meanings, and showed that the number of meanings of a word grows as the square root of its frequency. He derived this relationship from two assumptions: that words follow Zipf's law for word frequencies (a power law dependency between frequency and rank) and Zipf's law of meaning distribution (a power law dependency between number of meanings and rank). Here we show that a single assumption on the joint probability of a word and a meaning suffices to infer Zipf's meaning-frequency law or relaxed versions. Interestingly, this assumption can be justified as the outcome of a biased random walk in the process of mental exploration.   
\end{abstract}

\section*{Introduction}



G. K. \citeA{Zipf1949a} investigated many statistical regularities of language. Some of them have been investigated intensively such as Zipf's law for word frequencies \cite{Fedorowicz1982a, Ferrer2009a, Font2013a, Ferrer2016b} or Zipf's law of abbreviation \cite{Strauss2006a, Ferrer2012d}. 
Some others, such as Zipf's law of meaning distribution have received less attention.  
In his pioneering research, \citeA{Zipf1945a} found that more frequent words tend to have more meanings. The functional dependency between $\mu$, the number of meanings of a word, and $f$, the frequency of a word, has been approximated with \cite{Ilgen2007a, Baayen2005a, Zipf1945a}
\begin{equation}
\mu \propto f^{\delta},
\label{meaning_frequency_law_equation}
\end{equation}
where $\delta$ is a constant such that $\delta \approx 1/2$. Eq. \ref{meaning_frequency_law_equation} defines Zipf's meaning-frequency law.
Equivalently, the meaning-frequency law can be defined as
\begin{equation}
f \propto \mu^{1/\delta}.
\label{mirror_meaning_frequency_law_equation}
\end{equation}
G. K. Zipf derived the meaning-frequency law assuming two laws, the popular Zipf's law for word frequencies and the law of meaning distribution. On the one hand, Zipf's law for word frequencies states the relationship between the frequency of a word and its rank (the most frequent word has rank $i=1$, the 2nd most frequent word has rank $i=2$ and so on) as  
\begin{equation}
f \propto i^{-\alpha},
\label{frequency_rank_law_equation}
\end{equation}
where $\alpha \approx 1$ is a constant \cite{Zipf1945a,Zipf1949a}. On the other hand, the law of meaning distribution \cite{Zipf1945a,Zipf1949a} states that
\begin{equation}
\mu \propto i^{-\gamma},
\label{law_of_meaning_distribution_equation}
\end{equation}
where $\gamma \approx 1/2$. Notice that $i$ is still the rank of a word according to its frequency.
The constants $\alpha$, $\gamma$ and $\delta$ can be estimated applying some regression method as in Zipf's pioneering research \cite{Zipf1945a,Zipf1949a}.

Sometimes, power-laws such as those described in Eqs. \ref{meaning_frequency_law_equation}-\ref{law_of_meaning_distribution_equation} are defined using asymptotic notation. For instance, random typing yields $f = \Theta(i^{-\alpha})$, {\em i.e.} for sufficiently large $i$, one has that \cite{Conrad2004a} 
\begin{equation}
a_1 i^{-\alpha} \leq f \leq a_2 i^{-\alpha},
\label{frequency_rank_law_in_random_typing_equation}
\end{equation}
where $a_1$ and $a_2$ are constants such that $a_1 \leq a_2$. Eq. \ref{frequency_rank_law_in_random_typing_equation} can be seen as relaxation of Eq. \ref{frequency_rank_law_equation}. Similarly, Heaps' law on the relationship between $V$, the number of types, as function of $T$, the number of tokens, is defined as $V = \Theta(T^\beta)$ with $0 < \beta < 1$ \cite{Baeza_Yates2000a}, a relaxed version of $V \propto T^\beta$ (see \citeA{Font2015a} for a critical revision of the power law model for Heaps' law). 

The meaning-frequency law (Eq. \ref{meaning_frequency_law_equation}) and the law of meaning distribution (Eq. \ref{law_of_meaning_distribution_equation}) predict the number of meanings of a word using different variables as predictors. The meaning-frequency law has been confirmed empirically in various languages: directly through Eq. \ref{meaning_frequency_law_equation} in Dutch and English \cite{Baayen2005a} or indirectly through Eq. \ref{law_of_meaning_distribution_equation} and the assumption of Zipf's law \cite{Zipf1945a, Ilgen2007a} in Turkish and English. Qualitatively, 
the meaning-frequency law defines a positive correlation between frequency and the number of meanings. Using a proxy of word meaning the qualitative version of the law has been found in dolphin whistles \cite{Ferrer2009f} and in chimpanzee gestures \cite{Hobaiter2014a}. Thus, the law is a candidate for a universal property of communication.

\citeA{Zipf1945a} argued that Eq. \ref{meaning_frequency_law_equation} with $\delta= 1/2$ follows from Eq. \ref{frequency_rank_law_equation} with $\alpha = 1$ and Eq. \ref{law_of_meaning_distribution_equation} with $\gamma = 1/2$. Indeed, it has been proven that Eq. \ref{meaning_frequency_law_equation} with $\delta = \gamma/\alpha$ follows from Eqs. \ref{meaning_frequency_law_equation} and \ref{law_of_meaning_distribution_equation} \cite{Ferrer2014d}.  Here we consider alternative derivations of Eq. 
\ref{meaning_frequency_law_equation} or relaxed versions of Eq. 
\ref{meaning_frequency_law_equation} from the assumption of a biased random walk \cite{Sinatra2011a,Gomez-Gardenes2008a} over word-meaning associations. The remainder of the article is organized as follows. 

First, we will present the mathematical framework. 

Second, we will present a minimalist derivation of the meaning-frequency law (Eq. \ref{meaning_frequency_law_equation}) with $\delta= 1/2$ law that is based on just one assumption on the joint probability of a word and a meaning. Suppose a word $s$ that is connected to $\mu$ meanings and a meaning $r$ that is connected to $\omega$ words.  Assuming that the joint probability of $s$ and $r$ is proportional to $\mu$ if $s$ and $r$ are connected and zero otherwise, suffices to obtain Eq. \ref{meaning_frequency_law_equation} with $\delta = 1/2$. A problem of the argument is that the definition is somewhat arbitrary and theoretically superficial. 

Third, we will replace 
this simplistic assumption by a more fundamental assumption, namely that the joint probability of $s$ and $r$ is proportional to $\mu\omega$ if $s$ and $r$ are connected and zero otherwise. This assumption is a more elegant solution for two reasons: it corrects the arbitrariness of the assumption of the minimalist derivation, fits into standard network theory and it can be embedded into a general theory of communication. From this deeper assumption we derive the meaning-frequency law following three major paths. The 1st path consists of assuming the principle of contrast \cite{Clark1987a} or the principle of no synonymy \cite[p. 67]{Goldberg1995}, namely, $\omega \leq 1$ for all words, which leads exactly to Eq. \ref{meaning_frequency_law_equation}. The 2nd path consists of assuming that meaning degrees are mean independent of word degrees, which leads to a mirror of the meaning-frequency law (Eq. \ref{mirror_meaning_frequency_law_equation}) 
\begin{equation}
E[p|\mu] \propto \mu^{1/\delta}, 
\label{mean_independence_law_equation}
\end{equation}
where $E[p|\mu]$ is the expectation of $p$, the probability of a word, knowing that its degree is $\mu$. Notice that $p$ is linked with $f$ as $p \approx f/L$, where $L$ is the length of a text in tokens, {\em i.e.} total number of tokens (the sum of all type frequencies) \cite{Moreno2016a}. 
The 3rd path makes no assumption to obtain a relaxed version of the meaning-frequency law, namely the number of meanings is bounded above and below by two power-laws over $f$, {\em i.e.} 
\begin{equation*}
b_1 f^\delta \leq \mu \leq b_2 f^\delta,
\end{equation*}
where $b_1$ and $b_2$ are constants such that $b_1 \leq b_2$. The result can be summarized as 
\begin{equation}
\mu =  \Theta(f^\delta).
\label{asymptotic_meaning_frequency_law_equation}
\end{equation}
Put together, these three paths strongly suggest that languages are channeled to reproduce Zipf's meaning-frequency law.

Fourth, we will review 
a family of optimization models of communication that was put forward to investigate the origins of Zipf's law for word frequencies  \cite{Ferrer2007a} but that has recently been used to shed light on patterns of vocabulary learning and the mapping of words into meanings \cite{Ferrer2013g}. Interestingly, models from that family give Eq. 
\ref{meaning_frequency_law_equation} with $\delta = 1$ \cite{Ferrer2014d}. Crucially, however, the true exponent is $\delta \approx 1/2$ \cite{Zipf1945a,Ilgen2007a}. The mismatch should not be surprising. Imagine that a speaker has to choose a word for a certain meaning. Those models assume that given a meaning, all the words with that meaning are equally likely \cite{Ferrer2007a}. However, this simple assumption is not supported by psycholinguistic research \cite{Snodgrass1980a}. 
We will show
how to modify their definition so that the words that are used for a certain meaning do not need to be equally likely and one can obtain Eq. \ref{meaning_frequency_law_equation} with $\delta = 1/2$ or relaxed versions.  
Finally, we will discuss the results, highlighting the connection with biased random walks, and indicate 
directions for future research.

\section*{A mathematical framework}

As in the family of models of communication above, we assume a repertoire of $n$ words, $s_1$,...,$s_i$,...$s_n$ and a repertoire of $m$ meanings, $r_1$,...,$r_i$,...,$r_m$. Words and meanings are associated through an $n \times m$ adjacency matrix $A = \{a_{ij}\}$: $a_{ij} = 1$ if $s_i$ and $r_j$ are associated ($a_{ij} = 0$ otherwise). $A$ defines the edges of an undirected bipartite network of word-meaning associations. The degree of the $i$-th word is
\begin{equation}
\mu_i = \sum_{j=1}^m a_{ij} 	
\label{word_degree_equation}
\end{equation}
while the degree of the $j$-th meaning is
\begin{equation}
\omega_j = \sum_{i=1}^n a_{ij}. 
\label{meaning_degree_equation}
\end{equation}

In human language, the relationship between sound and meaning has been argued to be arbitrary to a large extent \cite{Saussure1916a, Hockett1966a, Pinker1999a}. That is, there is no intrinsic relationship between the word form and its meaning.  For example the word "car" is nothing like an actual automobile. An obvious exception are onomatopoeias, which are relatively rare in language. However, despite the immense flexibility of the world's languages, some sound-meaning associations are preferred by culturally, historically, and geographically diverse human groups \cite{Blasi2016a}. The framework above is agnostic concerning the type of association between sound and meaning. By doing that, we are borrowing the abstract perspective of network theory, that is {\em a priori} neutral concerning the nature or the origins of the edges \cite{Newman2010a,Barthelemy2011a}. Our framework could be generalized to accommodate Peirce's classic types of reference, {\em i.e.}, iconic, indexical and symbolic \cite{Deacon1997}, or the state-of-the-art on the iconicity/systematicity distinction \cite{Dingemanse2015a}. An crucial reason to remain neutral is that the distinctions above were not made when defining the laws of meaning that are the target of this article.

The framework allows one to model lexical ambiguity: a lexically ambiguous word is a word such that its degree is greater than one. Although the model starts from a flat hierarchy of concepts (by default all concepts have the same degree of generality), a word with an abstract meaning could be approximated either as a word linked to a single abstract concept or as a word linked to the multiple specific meanings it covers \cite{Ferrer2013g}. As for the latter approach, the word for vehicle would be linked to the meanings for car, bike, ship, airplane,...

Suppose that $p(s_i, r_j)$ is the joint probability of the unordered pair formed by $s_i$ and $r_j$. 
By definition, 
\begin{equation}
\sum_{i=1}^n \sum_{j=1}^m p(s_i, r_j) = 1.
\label{normalization_of_joint_probability_equation}
\end{equation}
The probability of $s_i$ is   
\begin{equation}
p(s_i) = \sum_{j=1}^m p(s_i, r_j)
\label{word_probability_equation}
\end{equation}
and the probability of $r_i$ is   
\begin{equation}
p(r_j) = \sum_{i=1}^n p(s_i, r_j).
\label{meaning_probability_equation}
\end{equation}
Our model shares the assumptions of distributional semantics that the meaning of a word is represented as a vector of the weights of different concepts for that word \cite{Lund1996a}. In our framework, the meaning of the word $s_i$, is represented by the $m$-dimensional vector
\begin{equation*}
\{p(s_i, r_1),...,p(s_i, r_j),...,p(s_i, r_m)\}
\end{equation*}

The joint probabilities $p(s_i, r_j)$ for all words and meanings defines a weighted matrix of the same size of $A$. 
In the coming sections, we will derive the meaning frequency-law defining $p(s_i, r_j)$ as a function of $A$.
Put differently, we will derive the law from a weighted undirected bipartite graph that is build from the unweighted undirected graph defined by $A$. This organization in two graphs (one unweighted and the other weighted) instead of a single weighted graph is borrowed from successful models of communication \cite{Ferrer2013g} and allows one to apply the theory of random walks \cite{Sinatra2011a,Gomez-Gardenes2008a} as we will see later on.

\section*{A minimalist derivation of the law}

The law of meaning distribution can be derived by making just one rather simple assumption, {\em i.e.}
\begin{equation}
p(s_i, r_j) \propto a_{ij}\mu_i,
\label{minimalistic_joint_probability_seed_equation}
\end{equation}
Applying Eq. \ref{normalization_of_joint_probability_equation}, one obtains 
\begin{equation}
p(s_i, r_j) = c a_{ij}\mu_i,
\label{minimalistic_joint_probability_equation}
\end{equation}
where $c$ is a normalization constant defined as
\begin{equation*}
c = \frac{1}{\sum_{i=1}^n \sum_{j=1}^m a_{ij}\mu_i} = \frac{1}{\sum_{i=1}^n \mu_i \sum_{j=1}^m a_{ij}} = \frac{1}{\sum_{i=1}^n \mu_i^2}.
\end{equation*}
Notice that $c$ is not a parameter and its value if determined by the definition of probability in Eq. \ref{normalization_of_joint_probability_equation}.
Applying Eq. \ref{word_probability_equation} to Eq. \ref{minimalistic_joint_probability_equation} gives 
\begin{equation*}
p(s_i) = c \mu_i^2,
\end{equation*}
namely Eq. \ref{meaning_frequency_law_equation} with $\delta = 1/2$.
Our derivation of the strong and relaxed version of the meaning-frequency law is simpler than that of Zipf's in the sense that it requires assuming a smaller number of equations (we are assuming only Eq. \ref{minimalistic_joint_probability_seed_equation} while Zipf assumed Eqs. \ref{frequency_rank_law_equation} and \ref{law_of_meaning_distribution_equation}). However, the challenge of our approach is the justification of Eq. \ref{minimalistic_joint_probability_seed_equation}. 

\section*{A theoretical derivation of the law}

The definition of $p(s_i, r_j)$ in Eq. \ref{minimalistic_joint_probability_seed_equation} suffices as a model but not for the construction of a real theory of language. Eq. \ref{minimalistic_joint_probability_seed_equation} is simple but somewhat arbitrary: the degree of the word, $\mu_i$, contributes raised to 1 but the degree of the meaning, $\omega_j$ has no direct contribution, or one may say that it contributes raised to 0. Therefore, a less arbitrary equation would be  
\begin{equation}
p(s_i, r_j) \propto a_{ij} (\mu_i \omega_j)^\phi.
\label{balanced_joint_probability_seed_equation}
\end{equation}
where $\phi$ is a positive parameter ($\phi \geq 0$).  
Applying Eq. \ref{normalization_of_joint_probability_equation} to Eq. \ref{balanced_joint_probability_seed_equation}, one obtains
\begin{equation}
p(s_i, r_j) = c a_{ij} (\mu_i \omega_j)^\phi.
\label{balanced_joint_probability_equation}
\end{equation}
with $c = 1/M$ and 
\begin{equation}
M = \sum_{i=1}^n \sum_{j=1}^m a_{ij} (\mu_i \omega_j)^\phi = \sum_{i=1}^n \mu_i^\phi \sum_{j=1}^m a_{ij} \omega_j^\phi.
\label{total_equation}
\end{equation}
Notice that $\phi$ is the only parameter of the model given $n$ and $m$.
Applying Eq. \ref{word_probability_equation} to Eq. \ref{balanced_joint_probability_equation}, one obtains
\begin{equation}
p(s_i) = c \mu_i^\phi \sum_{j=1}^m a_{ij}\omega_j^\phi. 
\label{new_word_probability_equation}
\end{equation}

Eq. \ref{balanced_joint_probability_seed_equation} is theoretically appealing for various reasons. If $p(s_i, r_j)$ is regarded as the weight of the association between $s_i$ and $r_j$, it defines the general form of the relationship between the weight of and edge and the product of the degrees of vertices at both ends that is found in real networks \cite{Barrat2004a}. 
For this reason, a unipartite version of Eq. \ref{balanced_joint_probability_seed_equation} is assumed to study dynamics on networks \cite{Baronchelli2011a}.   
When $\phi = 0$, it matches the definition of models about the origins of Zipf's law for word frequencies \cite{Ferrer2004e}, the variation of the exponent of the law \cite{Ferrer2004a,Ferrer2005e} and vocabulary learning \cite{Ferrer2013g}. When $\phi = 1$, it defines an approximation to the stationary probability of observing a transition involving $s_i$ and $r_j$ in a random walk on a network that is biased to maximize the entropy rate of the walks (Appendix \ref{random_walks_appendix}), thus suggesting that the meaning-frequency law 
could be a manifestation of a particular random walk process on semantic memory.

Two equivalent linguistic principles, the principle of contrast \cite{Clark1987a} and the principle of no synonymy \cite[p. 67]{Goldberg1995} can be implemented in our model as $\omega_j \in \{0,1\}$. From an algebraic standpoint, the condition $\omega_j \in \{0,1\}$ is equivalent to orthogonality of the word vectors of the matrix $A$. If $A_{i*}$ indicates the row vector of $A$ for the $i$-th word, $A_{i*}$ and $A_{k*}$ are orthogonal if and only if $A_{k*} \cdot A_{k*}=0$, where the dot indicates the scalar product of two vectors. To simplify matters, we assume that there is no row vector of $A$ that equals $\vec{0}$, a vector that has $0$ in all components. So far we have used $\mu_i$ and $\omega_j$ to refer, respectively, to the degree of the $i$-th word and the $j$-th meaning. We define $\mu_i^e$ and $\omega_i^e$ as the degree of the word and the degree of the meaning of the $i$-th edge. $\mu_i^e$ and $\omega_i^e$ are components of the vectors $\vec{\mu_i^e}$ and $\vec{\omega_i^e}$, respectively. We have $\vec{\mu_i^e} \cdot \vec{\omega_i^e} >0$ because $\mu_i^e, \omega_i^e \geq 1$ by definition. A deeper insight can be obtained with the concept of remaining degree, the degree at one end of the edge after subtracting the unit contribution of the edge \cite{Newman2002a}. The vectors of remaining degrees are then
\begin{eqnarray*}
\vec{\mu_i'^e} = \vec{\mu_i^e} - \vec{1}\\
\vec{\omega_i'^e} = \vec{\omega_i^e} - \vec{1}.
\end{eqnarray*}
The condition $\omega_j \in \{0,1\}$ is equivalent to $\vec{\omega_i'^e} = \vec{0}$. $\omega_j \in \{0,1\}$ leads to $\vec{\mu_i'^e}\vec{\omega_i'^e} = 0$ but trivially for being $\vec{\omega_i'^e}$ null.

The assumption $\phi = 1$ and $\omega_j \in \{0,1\}$ (orthogonality of row vectors of $A$), transform Eq. \ref{new_word_probability_equation} into Eq. \ref{minimalistic_joint_probability_seed_equation} because $a_{ij} = 0$ and $\omega_j=0$ are equivalent when $\omega_j$ does not exceed 1. In general, Eq. \ref{new_word_probability_equation} combined with the principle of contrast gives 
\begin{eqnarray*}
p(s_i) & = & c \mu_i^\phi \sum_{j=1}^m a_{ij} \\
       & = & c \mu_i^{\phi+1} 
\end{eqnarray*}
and finally Eq. \ref{meaning_frequency_law_equation} with 
\begin{equation*}
\delta=\frac{1}{\phi + 1}.
\end{equation*}
When $\phi = 1$, we get $\delta=1/2$ again.
Interestingly, the principle of contrast follows from the principle of mutual information maximization, a more fundamental principle that allows one to predict vocabulary learning in children and that can be combined with the principle of entropy minimization to predict Zipf's law for word frequencies \cite{Ferrer2013g}.  
With Eq. \ref{balanced_joint_probability_seed_equation}, we follow Bunge \cite[pp. 32-33]{Bunge2013a} preventing scientific knowledge from becoming ``an aggregation of disconnected information'' and aspiring to build a ``system of ideas that are logically connected among themselves''. 
  
It is possible to obtain a relaxed meaning frequency-law under more general conditions. In particular, we would like to get rid of the heavy constraint that meaning degrees cannot exceed one. Suppose that $d$ is a constant such that $0 < d \leq n$. Some obvious but not very general conditions are $\omega_j \in \{0,d\}$ for all $j$ or $\omega_j = d$ for all $j$. It is easy to see that they lead again to Eq. \ref{minimalistic_joint_probability_seed_equation} when $\phi = 1$.  
A more general condition can be defined as follows. First, we define $E[\omega^\phi | \mu]$ as the conditional expectation of $\omega^\phi$ given $\mu$ for an edge. Here $\mu$ and $\omega$ are the degrees at both ends of an edge. Then suppose that $A$ is given and that 
the $E[\omega^\phi | \mu] = E[\omega^\phi]$
$\omega^\phi$ is mean independent of $\mu$, namely $E[\omega^\phi | \mu] = E[\omega^\phi]$ for each value of $\mu$ \cite{Kolmogorov1956a,Poirier1995a}, then the expectation of $p(s_i)$ (as defined in Eq. \ref{new_word_probability_equation}) knowing $\mu_i$ is
\begin{eqnarray*}
E[p(s_i)|\mu_i] & = & E\left[c \mu_i^\phi \sum_{j=1}^m a_{ij}\omega_j^\phi \middle|  \mu_i \right] \\ 
                & = & c \mu_i^\phi E\left[\sum_{j=1}^m a_{ij}\omega_j^\phi \middle| \mu_i \right] \\
                & = & c \mu_i^\phi \sum_{j=1}^m a_{ij} E\left[\omega_j^\phi \middle| \mu_i \right] \\
                & = & c \mu_i^\phi \sum_{j=1}^m a_{ij} E\left[\omega_j^\phi \right] \\
                & = & c E[\omega_j] \mu_i^{\phi + 1},
\end{eqnarray*}
which can be seen as a regression model \cite{Ritz2008a} for the meaning-frequency law (Eq. \ref{meaning_frequency_law_equation}) with word degree as predictor. 
Notice that mean independence is a more general condition than mutual or statistical independence but a particular case of uncorrelation \cite{Ferrer2012h}. 

So far, we have seen ways of obtaining the meaning-frequency law from Eq. \ref{balanced_joint_probability_seed_equation} making further assumptions. It is possible to obtain a relaxed version of the meaning-frequency law making no additional assumption (except Eq. \ref{balanced_joint_probability_seed_equation} or the biased random walk that justifies it). 
Eq. \ref{new_word_probability_equation} can be expressed as 
\begin{equation}
p(s_i) = \mu_i^\phi \sum_{j=1}^m a_{ij} T_j 
\label{temporary_new_word_probability_equation}
\end{equation}
with 
\begin{equation*}
T_j = c \omega_j^\phi.
\end{equation*}
Assuming that 
\begin{equation*}
T_{min} \leq T_j \leq T_{max},
\end{equation*}
Eq. \ref{temporary_new_word_probability_equation} leads to 
\begin{equation}
T_{min} \mu_i^{\phi + 1} \leq p(s_i) \leq T_{max} \mu_i^{\phi+1}  
\label{relaxed_mirror_meaning_frequency_law_equation}
\end{equation} 
or equivalently
\begin{equation*}
\frac{1}{T_{min}} p(s_i)^{\frac{1}{\phi + 1}} \leq \mu_i \leq \frac{1}{T_{max}} p(s_i)^{\frac{1}{\phi + 1}}.
\end{equation*} 
Recalling $p \approx f/L$, these results can be summarized using asymptotic notation as $f = \Theta(\mu^{\phi+1})$ or $\mu = \Theta(f^{1/(\phi + 1)})$.
The power of the bounds above depends on the gap between $T_{min}$ and $T_{max}$. The gap can be measured with the ratio 
\begin{equation*}
\frac{T_{max}}{T_{min}} = \frac{\omega_{max}^\phi}{\omega_{min}^\phi}, 
\end{equation*}
where $\omega_{min}$ and $\omega_{max}$ are the minimum and the maximum meaning degree, respectively.
The principle of mutual information maximization between words and meanings, a general principle of communication \cite{Ferrer2013g}, puts pressure for concordance with the meaning-frequency law. To see it, we consider two cases: $n\leq m$ and $m \leq n$. When $n \leq m$, its maximization predicts $\omega_j \leq 1$ (Appendix \ref{mutual_information_maximixation_appendix}). As unlinked meanings are irrelevant (they do not belong to the support set), we have $\omega_{min} = 1$. As pressure for mutual information maximization increases, $\omega_{max}$ tends to 1 and thus $T_{max}/T_{min}$ tends to $1$. Put differently, the gap between the upper and the lower bound in Eq. \ref{relaxed_mirror_meaning_frequency_law_equation} reduces as pressure for mutual information maximization increases. When $n \geq m$, mutual information maximization predicts that $\omega_j = d$, where $d$ is an integer such that $d \in [1, \lfloor n/m \rfloor]$ (Appendix \ref{mutual_information_maximixation_appendix}). We have seen above that one obtains the meaning-frequency law (Eq. \ref{meaning_frequency_law_equation}) immediately from Eq. \ref{balanced_joint_probability_seed_equation} when $\omega_j$ is constant. We conclude that the chance of observing the meaning-frequency law increases as pressure for mutual information maximization increases.    

\section*{A family of optimization models of communication}

Here we revisit a family of optimization models of communication \cite{Ferrer2007a} in light of the results of the previous sections. 
These models share the assumption that the probability that a word $s_i$ is employed to refer to meaning $r_j$ is proportional to $a_{ij}$, {\em i.e.}
\begin{equation}
p(s_i | r_j) \propto a_{ij},
\label{conditional_word_probability_seed_equation}
\end{equation}
Applying 
\begin{equation*}
\sum_{i=1}^n p(s_i | r_j) = 1
\end{equation*}
to Eq. \ref{conditional_word_probability_seed_equation}, we obtain
\begin{equation}
p(s_i | r_j) = \frac{a_{ij}}{\omega_j}.
\label{conditional_word_probability_equation}
\end{equation}
We adopt the convention $p(s_i|r_j) = 0$ when $\omega_j = 0$.   

Eq. \ref{conditional_word_probability_equation} defines the probability of transition of a standard (unbiased) random walk to a word \cite{Noh2004a}, {\em i.e.}
given a meaning, all related words are equally likely. This is unrealistic in light of picture naming norms \cite{Snodgrass1980a,Dunabeitia2017a}. Consider the picture-naming norms compiled by \citeA{Snodgrass1980a}, who simply asked participants to name 260 black-and-white line drawings of common objects. For some objects ({\em e.g.}, balloon, banana, sock, star) there was 100\% agreement among the participants for the word used to name the pictured object. However, for other objects there was considerable variability in the word used to name the pictured object. Important for the present argument, the other words that were used in such cases were not selected with equal likelihood. For example, the picture of a {\em wineglass} had 50\% agreement, with the word {\em glass} (36\% of the responses) and the word {\em goblet} (14\% of the responses) also being used to name the object, showing that all the words that could be used for a given meaning are not equally likely. 
Although subjects tend to provide more specific responses when the concept is presented in textual form with respect to a visual form presentation \cite{Tversky1983a}, 
we used the visual case simply to challenge the assumption of an unbiased random walk in general and justify a more realistic approach.

In contrast to Eq. \ref{conditional_word_probability_equation}, the fundamental assumption in Eq. \ref{balanced_joint_probability_seed_equation} leads to 
\begin{equation}
p(s_i | r_j) = \frac{a_{ij}\mu_i^\phi}{\sum_{k}a_{kj}\mu_k^\phi},
\label{biased_conditional_word_probability_equation}
\end{equation}
namely the transition probabilities of a biased random walk when $\phi > 0$ \cite{Sinatra2011a,Gomez-Gardenes2008a}. To see it, notice that the combination of Eq. \ref{meaning_probability_equation} and \ref{balanced_joint_probability_equation} produces 
\begin{equation}
p(r_j) = \sum_{i=1}^n p(s_i, r_j) = c \omega_j^\phi \sum_{i=1}^n a_{ij}\mu_i^\phi.
\label{biased_meaning_probability_equation}
\end{equation}
Recalling the definition of conditional probability 
\begin{equation*}
p(s_i | r_j) = \frac{p(s_i, r_j)}{p(r_j)}
\end{equation*}
and applying Eq. \ref{balanced_joint_probability_equation} again, one obtains Eq. \ref{biased_conditional_word_probability_equation}. 

Recalling the definition of $\omega_j$ in Eq. \ref{meaning_degree_equation}, it is easy to realize that Eq. \ref{conditional_word_probability_equation} is a particular case of Eq. \ref{biased_conditional_word_probability_equation} with $\phi = 0$. While the family of models above stems from a concrete definition of a conditional probability, {\em i.e.} $p(s_i|r_j)$ in Eq. \ref{conditional_word_probability_equation}, the general model that we have presented in this article is specified by a definition of the joint probability, {\em i.e.} $p(s_i, r_j)$ in Eq. \ref{balanced_joint_probability_seed_equation}.


Models within that family are generated through
\begin{equation}
p(s_i, r_j) = p(s_i | r_j) p(r_j),
\label{model_generator_equation}
\end{equation}
assuming an unbiased random walk from a meaning to a word (Eq. \ref{conditional_word_probability_equation}) and making different assumptions on $p(r_j)$. 

If one assumes that all meanings are equally likely ($p(r_j)=1/m$ with $\omega_j \geq 1$) one obtains the 1st model  \cite{Ferrer2002a}. If one assumes that the probability of a meaning is proportional to its degree ($p(r_j) \propto \omega_j$) one obtains the 2nd model \cite{Ferrer2004e}. While in the 2nd model $p(r_j|s_i)$ defines an unbiased random walk from $s_i$ to $r_j$ (all $r_j$'s connected to $s_i$ are equally likely), this is not necessarily the case for the 1st model \cite{Ferrer2007a}. Therefore, the 2nd model defines a pure unbiased random walk while the 1st model is unbiased from meaning to words but biased from words to meanings. 

Now we will introduce a generalized version of the family of models above consisting of replacing 
Eq. \ref{conditional_word_probability_equation} by Eq. \ref{biased_conditional_word_probability_equation}
and generating the corresponding variants of the 1st and the 2nd model applying the same procedure as in the original family, namely via Eq. \ref{model_generator_equation}. Notice that Eq. \ref{biased_conditional_word_probability_equation} defines the probability of reaching $s_i
$ from $r_j$ in a biased random walk when $\phi > 0$. 

Concerning the 1st model, suppose that the probabilities of the meanings are given {\em a priori} (they are independent from the $A$ matrix), {\em e.g.}, all meanings are equally likely. Then it is easy to show that the model yields a relaxed version of the meaning frequency law, namely $\mu_i = \Theta(p(s_i)^\delta)$, the number of meanings is bounded above and below by two power-laws, {\em i.e.} (Appendix \ref{new_models_appendix})
\begin{equation}
b_1 p(s_i)^\delta \leq \mu_i \leq b_2 p(s_i)^\delta,
\label{bounds_on_degree_equation}
\end{equation}
where $b_1$ and $b_2$ are constants ($b_1 \leq b_2$) and $\delta = 1/(\phi + 1)$. Eq. \ref{bounds_on_degree_equation} defines non-trivial bounds when $\delta \neq 1$ (Appendix \ref{new_models_appendix}). The case $\delta = 1$ matches that 
an optimization model of Zipf's law for word frequencies \cite{Ferrer2004e, Ferrer2014d}.

To generate a variant of the 2nd model, recall that Eq. \ref{biased_conditional_word_probability_equation} comes from Eq. \ref{balanced_joint_probability_seed_equation}. 
Eqs. \ref{meaning_probability_equation} and \ref{balanced_joint_probability_equation} produce 
Eq. \ref{biased_meaning_probability_equation}.
This variant of the 2nd model derives all probability definitions from Eq. \ref{balanced_joint_probability_seed_equation}. 
We have shown above that this variant is able to generate the meaning-frequency law.

\section*{Discussion}

We have seen that it is possible to obtain the meaning-frequency law (Eqs. \ref{meaning_frequency_law_equation} and \ref{mirror_meaning_frequency_law_equation}) from Eq. \ref{balanced_joint_probability_seed_equation} making certain assumptions. We have also seen that a relaxed version of the law (Eq. \ref{asymptotic_meaning_frequency_law_equation} can be obtained from Eq. \ref{balanced_joint_probability_seed_equation} without making any further assumption.
Our findings suggest that word probabilities are channeled somehow to manifest the meaning-frequency law. We have seen that the principle of mutual information maximization contributes to the emergence of the law. Our derivation is theoretically appealing for various reasons. First, it is more parsimonious than G. K. Zipf's concerning the number of equations that are assumed (we only need Eq. \ref{balanced_joint_probability_seed_equation} while Zipf involved Eqs. \ref{frequency_rank_law_equation} and \ref{law_of_meaning_distribution_equation}). Second, it can help a family of optimization models of language to reproduce the meaning-frequency law. 


Therefore, a crucial assumption is Eq. \ref{balanced_joint_probability_equation}, that we have justified as the outcome of a random walk that is biased to maximize the entropy rate of the paths (Appendix \ref{random_walks_appendix}). A random walk is the correlate in network theory of the concept of mental exploration (navigation without a target or nonstrategic search) 
in cognitive science and related fields \cite{Baronchelli2013a}. Semantic memory
processes can be usefully theorized as searches over a network \cite{Thomson2014a, Abbot2015a} or some semantic space \cite{Smith2013a}. These approaches  support the hypothesis of a Markov chain process for memory search \cite{Bourgin2014a}, provide a deeper understanding of creativity \cite{Kenett2016a} and help to develop efficient navigation strategies \cite{Capitan2012a}.

A random walk in a unipartite word network of word-word associations has been argued to underlie Zipf's law for word frequencies \cite{Allegrini2003a}. 
Here we contribute with a new hypothesis linking random walks with a linguistic law: that the meaning-frequency law would be an epiphenomenon of a biased random walk over a bipartite network of word-meaning associations in the process of mental exploration. The bias consists of exploiting local information, namely the degrees of first neighbours \cite{Sinatra2011a}. Transitions to nodes with higher degree are preferred. Our model shows that it is possible to approximate the optimal solution to a problem (maximizing the entropy rate of the paths) following an apparently nonstrategic search \cite{Hills2012a,Abbot2015a}.

The probability of a word in Eq. \ref{new_word_probability_equation} defines the probability that a random walker visits the word in the long run. This probability is what the PageRank algorithm estimates in the context of a standard (non-biased) random walk \cite{Page1998a}. The assumption of a random walk with the particular bias above could help to improve random walk/PageRank methods to predict the prominence in memory of a word \cite{Griffiths2007a} or the importance of a tag \cite{Jaschke2007a}. 
A virtue of our biased random walk is that it predicts an uneven conditional probability of a word given a meaning (Eq. \ref{biased_conditional_word_probability_equation}) as it happens in real language \cite{Snodgrass1980a}. A standard (uniform) random walk cannot explain this fact and for that reason the family of optimization models of language revisited above fails to reproduce the meaning-frequency law with $\delta = 1/2$.    
 
Although biased random walks have already been used to solve information retrieval problems (see \citeA{Duarte2014a} and references therein), a bias based on the degree of neighbours has not been considered as far as we know. We hope that our results stimulate further research on linguistics laws and biased random walks in the information sciences. Specifically, we hope that our article becomes the fuel of future empirical research. 

\section*{Acknowledgements}
We are specially grateful to R. Pastor-Satorras and Massimo Stella for helpful comments and insights. We also thank S. Semple, M. Gentili and E. Bastrakova for helpful discussions. This research was supported by 
the grant APCOM (TIN2014-57226-P) from MINECO (Ministerio de Econom{\'i}a y Competitividad) and the grant 2014SGR 890 (MACDA) from AGAUR (Generalitat de Catalunya).

\appendix

\section{Random walks}

\label{random_walks_appendix}

We will show that Eq. \ref{balanced_joint_probability_seed_equation} defines the probability of observing a transition between $s_i$ and $r_j$ in any direction in a biased random walk. We will proceed in two steps. First, we will summarize some general results on biased random walks on unipartite networks and then we will adapt them to bipartite networks.
  
Suppose a unipartite network of $n$ nodes that is defined by an $n \times n$ adjacency matrix $B = \{b_{ij}\}$ such that 
$b_{ij} = 1$ if the $i$-th and the $j$-th node are connected and $b_{ij} = 1$ otherwise. Let 
$k_i$ be the degree of the $i$-th node, namely, 
\begin{equation*}
k_i = \sum_{j = 1}^n b_{ij}.
\end{equation*}
Suppose a random walk over the vertices of a network where $p(j | i)$ is the probability of jumping from $i$ to $j$. A first order approximation to the $p(j | i)$ that maximizes the entropy rate is \cite{Sinatra2011a}
\begin{equation}
p(j | i) = \frac{b_{ij} k_j}{\sum_{l=1}^n b_{il} k_l}.
\label{specific_unipartite_transition_probability_equation}
\end{equation}
We choose a generalization \cite{Gomez-Gardenes2008a} 
\begin{equation}
p(j | i) = \frac{b_{ij} k_j^\phi}{\sum_{l=1}^n b_{il} k_l^\phi},
\label{unipartite_transition_probability_equation}
\end{equation}
that gives Eq. \ref{specific_unipartite_transition_probability_equation} with $\phi = 1$.
The stationary probability of visiting the $i$-th vertex in the biased random walk defined by Eq. \ref{unipartite_transition_probability_equation} is \cite{Gomez-Gardenes2008a}
\begin{equation}
p(i) = \frac{k_i^\phi c_i}{T},
\label{uniparite_stationarty_probability_equation}
\end{equation}
where
\begin{equation}
c_i = \sum_{j=1}^n b_{ij}k_j^\phi
\label{unipartite_subweight_equation}
\end{equation} 
and
\begin{equation}
T = \sum_{i = 1}^n c_i k_i^\phi.
\label{unipartite_total_equation}
\end{equation}

Now we adapt the results above to a bipartite graph of word-meaning associations. As the graph is bipartite, the random walker will be alternating between words and meanings. The probability that the vertex visited is a word is $1/2$ (the same probability for a meaning).
Suppose that there are $n$ words and $m$ meanings. Recall that the bipartite network of word-meaning associations is defined by an $n \times m$ adjacency matrix $A = \{a_{ij}\}$ such that 
$a_{ij} = 1$ if the $i$-th word and the $j$-th meaning are connected and $a_{ij} = 1$ otherwise. 
$\mu_i$ is the degree of the $i$-th word is (Eq. \ref{word_degree_equation}) whereas $\omega_j$ is the degree of the $j$-th meaning (Eq. \ref{meaning_degree_equation}). 
The probability of jumping from $r_j$ to $s_i$ becomes (recall Eq. \ref{unipartite_transition_probability_equation}) 
\begin{equation*}
p_{v}(s_i |r_j) = \frac{a_{ij}\mu_i^\phi}{\sum_{l = 1}^n a_{lj}\mu_{l}^\phi}.
\end{equation*}
The probability of jumping from $s_i$ to $r_j$ is
\begin{equation}
p_{v}(r_j |s_i) = \frac{a_{ij}\omega_j^\phi}{\sum_{l = 1}^m a_{il}\omega_{l}^\phi}.
\label{conditional_meaning_probability_random_walk_equation}
\end{equation}

The stationary probability of visiting the word $s_i$ becomes (recall Eq. \ref{uniparite_stationarty_probability_equation} and \ref{unipartite_subweight_equation})
\begin{equation}
p_{v}(s_i) = \frac{\mu_i^\phi \sum_{j=1}^m a_{ij} \omega_j^\phi}{M_{v}},
\label{word_probability_random_walk_equation}
\end{equation}
where $M_v$ corresponds to $T$ in Eq. \ref{uniparite_stationarty_probability_equation}. Adapting Eqs. \ref{unipartite_total_equation} and \ref{unipartite_subweight_equation}, one obtains 
\begin{eqnarray*}
M_{v} & = & \sum_{i=1}^n \mu_i^\phi \sum_{j=1}^m a_{ij} \omega_j^\phi + \sum_{j=1}^m \omega_j^\phi \sum_{i=1}^n a_{ij} \mu_i^\phi \\
      & = & 2 M,
\end{eqnarray*}
where $M$ is defined as in Eq. \ref{total_equation}. Applying Eq. \ref{word_probability_random_walk_equation}, it is easy to see that
\begin{eqnarray*}
\sum_{i=1}^n p_{v}(s_i) & = & \frac{1}{2M} \sum_{i=1}^n \mu_i^\phi \sum_{j=1}^m a_{ij} \omega_j^\phi \\
                        & = & \frac{1}{2}.
\end{eqnarray*}  
as expected.

The combination of Eqs. \ref{conditional_meaning_probability_random_walk_equation} and \ref{word_probability_random_walk_equation} allows one to derive the probability of observing the transition from $s_i$ to $r_j$ as
%
\begin{eqnarray*}
p_{v}(s_i \rightarrow r_j) & = & p_{v}(r_j | s_i) p_{v}(s_i) \\
            & = & c_{v} a_{ij} (\mu_i \omega_j)^\phi,  
\end{eqnarray*}
where $c_{v} = 1/(2M)$. Similarly, the probability of observing the transition from $r_j$ to $s_i$ is 
\begin{eqnarray*}
p_{v}(s_i \leftarrow r_j) & = & p_{v}(s_i | r_j) p_{v}(r_j) \\
            & = & c_{v} a_{ij} (\mu_i \omega_j)^\phi.  
\end{eqnarray*}
Therefore the stationary probability of observing a transition between $s_i$ and $r_j$ in any direction (from $s_i$ to $r_j$ or from $r_j$ to $s_i$) is 
\begin{eqnarray*}
p(s_i, r_j) & = & p_{v}(s_i \rightarrow r_j) + p_{v}(s_i \leftarrow r_j) \\
            & = & 2c_v a_{ij} (\mu_i \omega_j)^\phi. \\
            & = & c a_{ij} (\mu_i \omega_j)^\phi.
\end{eqnarray*}
with $c=1/M$, as we wanted to show. 

Finally, we will link $p(s_i)$, the probability of a word that is used in the main text to derive the meaning-frequency law, with $p_v(s_i)$. Notice that $p(s_i) = p_v(s_i | S)$, the latter being the probability of visiting vertex $s_i$ knowing that it belongs to the partition $S$, the partition of words.
Since the graph is bipartite, $p_v(S)$, probability that the random walk is visiting a vertex of partition $S$, is $1/2$. The joint probability of visiting vertex $s_i$ and that the vertex belongs to $S$ is 
\begin{eqnarray*}
p_{v}(s_i, S) & = & p_v(S|s_i)p_v(s_i) \\
              & = & p_v(s_i).
\end{eqnarray*}
Therefore, 
\begin{eqnarray*}
p(s_i) & = & p_{v}(s_i | S) \\
       & = & \frac{p_{v}(s_i, S)}{p_{v}(S)} \\
       & = & 2 p_{v}(s_i).
\end{eqnarray*} 
Then $p(s_i)$ is the stationary probability of visiting $s_i$ in a biased random walk knowing that the vertex is in $S$. 



\section{Mutual information maximization} 
\label{mutual_information_maximixation_appendix}

Suppose that $I(S, R)$ is the mutual information between words ($S$) and meanings ($R$), that can be defined as 
\begin{equation}
I(S,R) = H(S) - H(S|R),
\label{mutual_information_equation}
\end{equation} 
where $H(S)$ is the entropy of words and $H(S|R)$ is the conditional entropy of words given meanings.
For the case $\phi = 0$, the configurations that maximize $I(S,R)$ when $n \leq m$ are defined by two conditions \cite{Ferrer2013g}
\begin{enumerate}
\item
$\mu_i = d$ with $d \in [1, \lfloor m/n \rfloor]$ for $i = 1,2,...,n$.  
\item
$\omega_j \in \{0, 1\}$ for $j = 1,2,...,m$.  
\end{enumerate}
When $n \geq m$, those configurations are the symmetric, {\em i.e.} \cite{Ferrer2013g}
\begin{enumerate}
\item
$\omega_j = d$ with $d \in [1,  \lfloor n/m \rfloor]$ for $j = 1,2,...,m$.  
\item
$\mu_i \in \{0, 1\}$ for $i = 1,2,...,n$.  
\end{enumerate}
Here we will show that the configurations that maximize $I(S, R)$ are the same as in the case $\phi = 0$ when $\phi$ is a positive and finite real number ($\phi \geq 0$). By symmetry, it suffices to show it for the case $n \leq m$. We will proceed in three steps. First, deriving the configurations minimizing $H(S|R)$.
Second, showing that the configurations above yield maximum $I(S,R)$. Third, showing that they are the only configurations.

Step 1: Recall that 
\begin{eqnarray*}
H(S|R) & = & E[H(S|r_j)] \\
       & = & \sum_{j=1}^m p(r_j) H(S|r_j)
\end{eqnarray*}
where $H(S|r_j)$ is the conditional entropy of words given the meaning $r_j$. 
Eq. \ref{biased_meaning_probability_equation} implies that $p(r_j) \neq 0$ is equivalent to $w_j > 0$ and then
\begin{eqnarray*}
H(S|R) & = & \sum_{\scriptsize \begin{array}{c} j=1 \\ p(r_j) \neq 0 \end{array}}^m p(r_j) H(S|r_j) \\
       & = & \sum_{\scriptsize \begin{array}{c} j=1 \\ w_j > 0  \end{array}}^m p(r_j) H(S|r_j).
\end{eqnarray*}
Knowing that
\begin{equation*} 
H(S|r_j) = -\sum_{i=1}^n p(s_i|r_j) \log p(s_i|r_j)
\end{equation*}
it is easy to see that $H(S|r_j)=0$ when $p(s_i|r_j) \in \{0,1\}$ for $i=1,2,...,n$: $0 \log 0 = 0$ by continuity since $x \log x \rightarrow 0$ as $x \rightarrow 0$ \cite[p. 14]{Cover2006a} and obviously $1 \log 1 = 0$. Eq. \ref{biased_conditional_word_probability_equation} implies that $p(s_i|r_j) = 1$ is equivalent to $s_i$ being the only neighbour of $r_j$, {\em i.e.} $\omega_j = 1$. Therefore, $H(S|R) = 0$ implies $\omega_j \leq 1$ for $j = 1,2,...,m$.  

Step 2: notice that the 2nd condition of the case $n \leq m$ above implies $H(S|R) = 0$ (recall Step 1). The 2nd condition transforms Eq. \ref{total_equation} into
\begin{equation*}
M = \sum_{i=1}^n \mu_i^\phi \sum_{j=1}^m a_{ij} = \sum_{i=1}^n \mu_i^{\phi+1}
\end{equation*}
and Eq. \ref{new_word_probability_equation} into 
\begin{equation*}
p(s_i) = c \mu_i^{\phi} \sum_{j=1}^m a_{ij} = c \mu_i^{\phi + 1}. 
\label{word_probability_with_contrast_equation}
\end{equation*}
Adding the 1st condition, one obtains
\begin{eqnarray*}
M = \sum_{i=1}^n d^{\phi+1} = n d^{\phi+1} \\
p(s_i) = c \mu_i^{\phi + 1} = \frac{1}{M} d^{\phi+1} = \frac{1}{n}. 
\end{eqnarray*}
and then $H(S) = \log n$ (as all words are equally likely).
Thus, $H(S)$ is taking its maximum possible value whereas $H(S|R)$ is taking its minimum value. As
$I(S,R) = H(S)-H(S|R)$, it follows that $I(S, R)$ is maximum. 

Step 3: notice that
\begin{itemize}
\item
If the 2nd condition fails, then $H(S|R) > 0$ and thus $I(S,R) < \log n$ even if $H(S)$ is maximum
because of Eq. \ref{mutual_information_equation}. Thus, the 2nd condition is required to maximize $I(S, R)$.
\item
If the 1st condition fails (while the 2nd condition holds), then words are not equally likely as the probability of a word is proportional to a power of its degree (Eq. \ref{word_probability_with_contrast_equation}). Then one has that $H(S) < \log n$ and it follows that $I(S,R)$ is not maximum because $I(S,R) \leq H(S)$.
\end{itemize}

\section{New models}
\label{new_models_appendix}

Combining Eqs. \ref{word_probability_equation} and \ref{biased_conditional_word_probability_equation}, one obtains
\begin{equation}
p(s_i) = \sum_{j=1}^{m} p(s_i, r_j) = \sum_{j=1}^m p(s_i|r_j)p(r_j) = \mu_i^\phi \sum_{j=1}^m a_{ij}T_j
\label{word_probability_in_new_model_equation}
\end{equation}
with
\begin{equation*}
T_j = \frac{p(r_j)}{\sum_{i=1}^n a_{ij} \mu_i^\phi}.
\end{equation*}
Suppose that 
\begin{equation*}
T_{min} \leq T_j \leq T_{max}
\end{equation*}
when $\omega_j > 0$. Eq. \ref{word_probability_in_new_model_equation} leads to  
\begin{equation*}
\mu_i^\phi \sum_{j=1}^m a_{ij} T_{min} \leq p(s_i) \leq \mu_i^\phi \sum_{j=1}^m a_{ij} T_{max}
\end{equation*}
and finally
\begin{equation}
T_{min} \mu_i^{\phi+1} \leq p(s_i) \leq T_{max} \mu_i^{\phi+1} 
\label{bounds_on_probability_new_model_equation}
\end{equation}
recalling the definition of $\mu_i$ in Eq. \ref{word_degree_equation}.
Equivalently, 
\begin{equation}
{T_{max}^{-\delta} p(s_i)^\delta \leq \mu_i \leq T_{min}^{-\delta} p(s_i)^\delta},
\label{bounds_on_degree_new_model_equation}
\end{equation}
with
\begin{equation*}
\delta = \frac{1}{\phi+1},
\end{equation*}
namely a relaxed version of the meaning-frequency law when $\phi = 1$. 

Notice that Eqs. \ref{bounds_on_probability_new_model_equation} and \ref{bounds_on_degree_new_model_equation} define non-trivial bounds in the sense that they are not expected from bounds on join-probability alone. If the range of variation of $p(s_i, r_j)$ satisfies
\begin{equation*}
\pi_{min} \leq p(s_i, r_j) \leq \pi_{max}
\end{equation*}
when $p(s_i, r_j)>0$, then Eq. \ref{word_probability_equation} gives
\begin{equation*}
\sum_{j=1}^m a_{ij} \pi_{min} \leq p(s_i) \leq \sum_{j=1}^m a_{ij} \pi_{max}
\end{equation*}
and then
\begin{equation*}
\pi_{min} \mu_i \leq p(s_i) \leq \pi_{max} \mu_i. 
\end{equation*}
Therefore, the finding that 
\begin{equation*}
b_1 p(s_i)^\delta \leq \mu_i \leq b_2 p(s_i)^\delta, 
\end{equation*}
where $b_1$ and $b_2$ are constants is trivial when $\delta = 1$. 

\bibliographystyle{apacite}
\bibliography{../biblio/rferrericancho,../biblio/complex,../biblio/ling,../biblio/cs,../biblio/cl,../biblio/maths} 

\end{document}